\documentclass[Afour,sageh,times]{sagej}

\usepackage{moreverb,url}
\usepackage{amsmath,amssymb,amsfonts}
\usepackage{multirow, array}
\usepackage{caption}
\usepackage[table]{xcolor}

\newcolumntype{P}[1]{>{\centering\arraybackslash}p{#1}}
\newcolumntype{M}[1]{>{\centering\arraybackslash}m{#1}}

\newcommand\BibTeX{{\rmfamily B\kern-.05em \textsc{i\kern-.025em b}\kern-.08em
T\kern-.1667em\lower.7ex\hbox{E}\kern-.125emX}}

\def\volumeyear{2025}
\def\journalname{Jounral Name}

\begin{document}

\runninghead{Mills}

\title{Scalable Cosmic AI Inference using Cloud Serverless Computing}

\author{Mills Staylor \textsuperscript{\rm 1},
     Amirreza Dolatpour Fathkouhi \textsuperscript{\rm 1},
     Md Khairul Islam \textsuperscript{\rm 1},
     Kaleigh O'Hara \textsuperscript{\rm 2},\\
     Ryan Ghiles Goudjil \textsuperscript{\rm 1}, 
     Geoffrey Fox  \textsuperscript{\rm 1},
     Judy Fox \textsuperscript{\rm 1,2}}

\affiliation{\textsuperscript{\rm 1}Department of Computer Science, University of Virginia\\ 
    \textsuperscript{\rm 2}School of Data Science, University of Virginia\\
    Charlottesville, Virginia, USA.}

\corrauth{Judy Fox}
\email{ckw9mp@virginia.edu}

\begin{abstract}
Large-scale astronomical image data processing and prediction are essential for astronomers, providing crucial insights into celestial objects, the universe’s history, and its evolution. While modern deep learning models offer high predictive accuracy, they often demand substantial computational resources, making them resource-intensive and limiting accessibility. We introduce the Cloud-based Astronomy Inference (CAI) framework to address these challenges. This scalable solution integrates pre-trained foundation models with serverless cloud infrastructure through a Function-as-a-Service (FaaS). CAI enables efficient and scalable inference on astronomical images without extensive hardware. Using a foundation model for redshift prediction as a case study, our extensive experiments cover user devices, HPC (High-Performance Computing) servers, and Cloud. Using redshift prediction with the AstroMAE model demonstrated CAI's scalability and efficiency, achieving inference on a 12.6 GB dataset in only 28 seconds compared to 140.8 seconds on HPC GPUs and 1793 seconds on HPC CPUs. CAI also achieved significantly higher throughput, reaching 18.04 billion bits per second (bps), and maintained near-constant inference times as data sizes increased, all at minimal computational cost (under \$5 per experiment). We also process large-scale data up to 1 TB to show CAI's effectiveness at scale. CAI thus provides a highly scalable, accessible, and cost-effective inference solution for the astronomy community.  The code is accessible at https://github.com/UVA-MLSys/AI-for-Astronomy.
\end{abstract}


\maketitle

\section{Introduction}
Astronomical images are vital to modern astrophysics, offering key insights into celestial objects, such as their shape \citep{galaxyzoo}, distance \citep{hubble}, and other fundamental characteristics that define our understanding of the universe. Large surveys like the Dark Energy Spectroscopic Instrument (DESI) \citep{desi} and the Sloan Digital Sky Survey (SDSS) \citep{sdss} provide extensive image datasets with unique attributes. For example, SDSS images consist of five spectral bands—u, g, r, i, and z—each focused on specific wavelengths: Ultraviolet (u) at 3543 Å, Green (g) at 4770 Å, Red (r) at 6231 Å, Near Infrared (i) at 7625 Å, and Infrared (z) at 9134 Å \citep{wavelengths}.

Deep learning foundation models trained on large-scale astronomical image datasets have proven to be powerful tools for improving the accuracy and efficiency of tasks such as redshift prediction, morphology classification, and similarity search \citep{hayat, astroclip, astromae}. Figure \ref{fig:model_parameters} highlights several notable deep-learning architectures that leverage astronomical images for prediction. For example, AstroCLIP \citep{astroclip} uses an image transformer with a 307M-parameter encoder, pretrained on approximately 197K images. AstroMAE \citep{astromae} combines frozen pretrained weights with fine-tunable parameters, totaling 10.4M learnable parameters. Similarly, \citep{henghes} trained their model on 1 million SDSS images using 7.8M parameters. In comparison, AstroMAE, trained on about 650K images, showed significant performance improvements over the \citep{henghes} model in redshift prediction. 

While these models yield impressive results, their large parameter counts require substantial computational resources, making both training and inference resource-intensive. The memory and computing resources on a standalone device become a limitation for users to run on large datasets and foundation models. There is a need for advanced infrastructure to overcome the limit of high-performance inference accessibility for many users \citep{neely}. Serverless computing is a recently popular Function-as-a-Service (FaaS) \citep{li2022serverless}, that lets developers write cloud functions in high-level languages (e.g. Python) and takes care of the complex infrastructure management itself.  

This study proposes a highly scalable serverless computing framework for using AWS \citep{serverless} to enhance the accessibility of a pretrained foundation model for astronomical images. This effectively reduces the computational demands on individual users. In summary, we offer the following contributions: 
\begin{itemize} 
    \item A novel Cloud-based Astronomy framework (named ``Cloud-based Astronomy Inference" (CAI)) to significantly enhance the scalability of foundation model inference on large astronomical images.
    \item Detailed experiments on the redshift prediction task using real-world galaxy images from the SDSS survey, comparing CAI's performance with other computing devices (e.g. personal, and HPC). 
    \item Our comprehensive performance analysis with inference time and throughput demonstrates that CAI effectively improves the scalability of foundation model inference in astronomy. 
\end{itemize}


\begin{figure*}[!ht]
  \centering
  \captionsetup{justification=centering}
  \includegraphics[width=.85\textwidth]{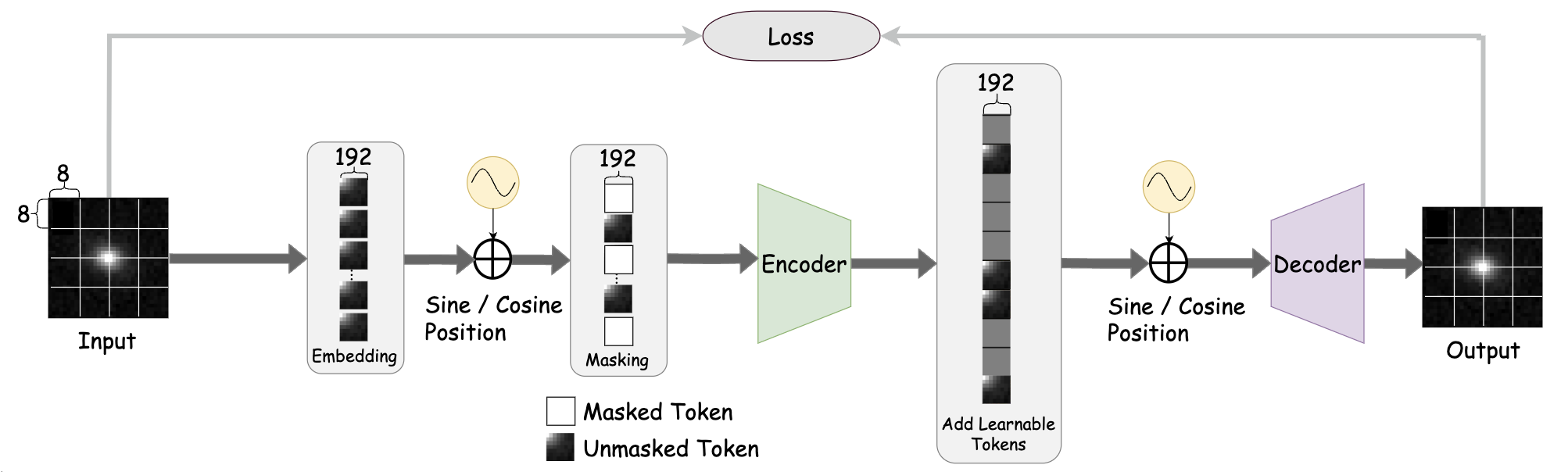}
  \caption{The architecture of masked autoencoder of AstroMAE \citep{astromae}.}
  \label{fig:masked_autoencoder}
\end{figure*}

\section{Related Work \label{sec:related_work}}

The immense size of astronomy datasets has made cloud services essential for efficiently storing and processing this data. \citep{faaique} highlighted the significant challenges such massive volumes pose, emphasizing cloud computing as a key solution to managing these issues. \citep{bigscience} investigated requirements for cloud computing used by scientists dealing with the increasing size of big science datasets. \citep{failuremanagement} accepted the invaluable nature of cloud computing for large astronomy datasets and used an astronomy case study to test their proposed cloud computing model that improves failure management. \citep{colitec_saas} argued that cloud services are nearly indispensable for managing astronomy datasets, especially those dense in metadata and images. In their study on variable star photometry, they implemented a Software-as-a-Service (SaaS) model to effectively address the substantial storage demands.

\citep{prado} leveraged cloud computing for the Mosaic tool, enabling the rapid and efficient creation of sky mosaics from images captured across different sky regions. \citep{kira} implemented the distributed astronomy image processing toolkit called Kira by running Apache Spark on an Amazon EC2 cloud. Their speed-up results and throughput indicate that this parallelized computing method is compatible with astronomy applications.  \citep{survey_astrobigdata} underscored the minimal scalability concerns of cloud services, noting the effectiveness of cloud-based distributed frameworks for tasks like redshift prediction. Furthermore, \citep{faas_radioastro} leveraged Function-as-a-Service (FaaS) models and decision-making systems to manage the computationally intensive processing of radio astronomy data. 

These studies highlight the essential role of cloud services in overcoming the computational and storage challenges inherent in modern astronomical research. To our knowledge, no prior work has leveraged serverless computing \citep{serverless} to enhance the scalability of foundation models for astronomical images.

\section{Problem Statement \label{sec:problem_statement}}

In astronomical imaging, much of the research has focused on enhancing deep learning model performance, with comparatively less attention to scaling inference capabilities. Although recent foundation models, trained on extensive astronomical image datasets, demonstrate versatility across various tasks, their high parameter count limits usability and scalability due to infrastructure constraints. To address this, a scalable framework is essential for efficient inference on large image volumes without added financial burdens. We introduce Cloud-based Astronomy Inference (CAI), which employs the Function-as-a-Service (FaaS) \citep{faas1} to enhance the scalability of foundation models trained on astronomical images. Based on our review, CAI is the first framework specifically designed to address the inference scalability of foundation models in astronomical imaging.

Although dark energy constitutes approximately 95\% of the universe's energy, our understanding of this mysterious force remains profoundly limited. Investigating and exploring its nature requires a large-scale collection of galaxy images, supported by advanced cosmological methods and theories \citep{jones2024redshift}. A cornerstone of these methods is the precise determination of a critical parameter: redshift. \textit{Redshift measures how much the light from a celestial object has been stretched}, providing crucial insights into the distances of these objects and the expansion of the universe \citep{hubble}.

In astronomical data analysis, the choice of computing infrastructure has a significant impact on efficiency, scalability, cost, and accessibility.  Personal computing devices, such as laptops and desktops, provide the most straightforward and accessible environment for initial exploration and analysis of limited scale. These devices have low upfront costs and minimal setup requirements, making them suitable for preliminary testing, educational purposes, or small datasets. However, limited memory and computational capabilities severely restrict their use for larger datasets or complex models. In our experiments, personal devices were unable to handle datasets larger than 8GB, underscoring their limited suitability for extensive astronomical data analysis.  High-performance computing (HPC) CPUs offer robust computational power and ample memory, enabling researchers to process large datasets effectively. These systems are reliable, capable of extensive batch processing, and suited to complex computational tasks that may not easily parallelize on GPUs.  GPUs within HPC environments significantly accelerate deep learning tasks due to their parallel processing architecture, which excels at inference workloads involving matrix operations typical of neural networks. HPC GPUs achieved much faster inference (140.8 s) compared to CPUs. GPUs are ideal for intensive model training and moderately large-scale inference tasks. However, GPU-based HPC systems come with high fixed and variable costs, as well as resource allocation complexities, making them less accessible to individual researchers or smaller institutions.

In this study, we evaluate the CAI scalability by applying it to the prediction of redshifts. For this purpose, we selected AstroMAE due to its superior performance and because it is pre-trained on a larger dataset.   Our proposed CAI framework leverages serverless cloud computing infrastructure to provide an optimal balance of scalability, cost-efficiency, and accessibility. CAI's linear scalability and minimal computational cost offer substantial advantages over traditional infrastructure. CAI eliminates upfront costs, complex maintenance, and resource management tasks associated with HPC clusters.

Future work will explore the scalability of additional models discussed in this context.

\section{Methodology \label{sec:methodology}}

We first collect the AstroMAE model pretrained on the large astronomy dataset. Then deploy it to our proposed cloud architecture for inference scalability benchmark.

\subsection{AstroMAE}

AstroMAE \citep{astromae} is a recent foundation model that captures general patterns in galaxy images for redshift prediction. It has two major phases:

\subsubsection{Pretraining:} 
Figure \ref{fig:masked_autoencoder} illustrates the pretraining process of AstroMAE's masked autoencoder \citep{devlin2018bert}. The masked autoencoder aims to reconstruct masked patches using unmasked ones. We mask 75\% of the embedded patches, the remaining 25\% are fed into the encoder. Initially, images are segmented into uniform patches of size 5×8×8 and embedded into 192-dimensional vectors with positional embedding. 

The reconstructed masked patches are compared to their original patches, enabling the model to learn meaningful representations. To promote learning of data patterns instead of memorizing patch positions, the embeddings are randomly shuffled before being input to the encoder. Compared to other pretraining methods like contrastive learning \citep{pmlr-v119-chen20j}, the masked autoencoder does not rely on specific augmentations, which can potentially increase the dataset size and computational demands. In contrast, AstroMAE's masked autoencoder processes 25\% patches, making it significantly more efficient. A crucial step for working on large astronomy data. AstroMAE also uses a modified ViT \citep{pcm_vit}, that contains a parallel convolutional module and performs even better. 

\subsubsection{Fine-tuning:}
During fine-tuning, the decoder is removed, leaving a frozen encoder that works with two additional models: a parallel Inception model and a magnitude block. The outputs from the frozen encoder, Inception model, and magnitude block are first processed through several layers individually, then concatenated and passed through additional layers for final processing. 

Let $V \in R^I$ represent the image data and $ O \in R^M $ represent the magnitude data, where \( I \) and \( M \) denote the dimensions of the image and magnitude data spaces, respectively.

A frozen pretrained encoder \( E: {R}^I \rightarrow {R}^C \) generates a latent space representation \( L_E \) from the image \( V \):

\begin{equation}
L_E = E(V).
\end{equation}

\begin{figure}[htb]
    \centering
    \captionsetup{justification=centering}
    \includegraphics[width=0.8\linewidth]{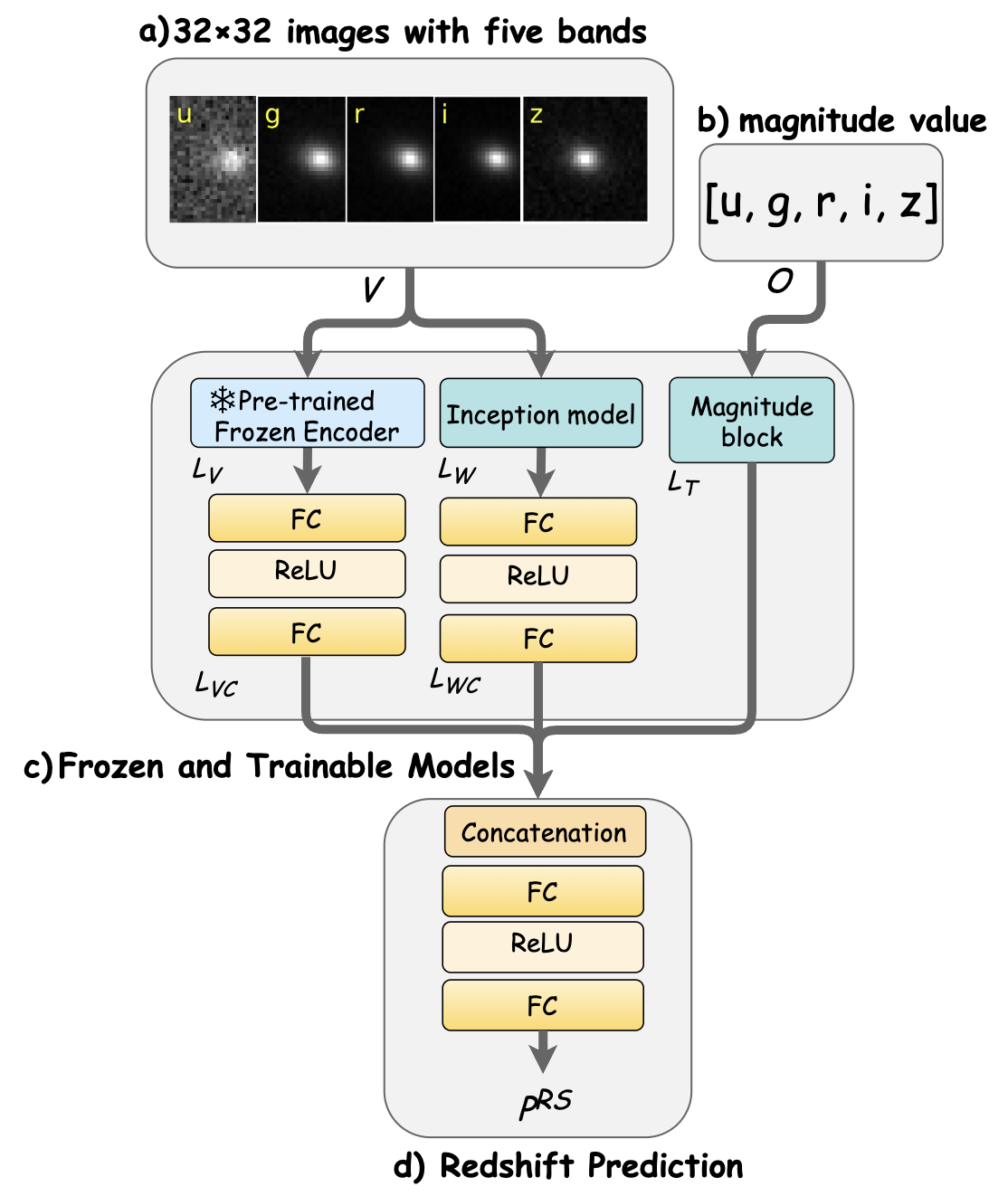}
    \caption{AstroMAE fine-tuning architecture.}
    \label{fig:astromae_architecture}
\end{figure}

Similarly, an inception model \( W: {R}^I \rightarrow {R}^Q \) extracts features \( L_W \) from the same image of Figure \ref{fig:astromae_architecture}a).

\begin{equation}
L_W = W(V).
\end{equation}

The magnitude data \( O \) from Figure \ref{fig:astromae_architecture}b) is processed through a magnitude block \( S: {R}^M \rightarrow {R}^T \), resulting in magnitude features \( L_T \):

\begin{equation}
L_T = S(O).
\end{equation}

To further process these image features, AstroMAE applies two fully connected layers with a ReLU activation in between to both \( L_E \) and \( L_W \). The resulting features are denoted as \( L_{EC} \) and \( L_{WC} \), representing the processed outputs of the frozen encoder and the inception model in Figure \ref{fig:astromae_architecture}c), respectively. 

\begin{align}
L_{EC} &= FC(\text{ReLU}(FC(L_E))), \\
L_{WC} &= FC(\text{ReLU}(FC(L_W))).
\end{align}

Finally, the redshift prediction \( P^{RS} \) is obtained as shown in Figure \ref{fig:astromae_architecture}d) by concatenating \( L_T \), \( L_{EC} \), and \( L_{WC} \), and then passing them through two fully connected layers with a ReLU activation function in between:

\begin{equation}
P^{RS} = FC(\text{ReLU}(FC(\text{Concat}(L_T, L_{EC}, L_{WC}))))
\end{equation}

The predicted redshift \( P^{RS} \) is compared with the actual redshift \( R^{RS} \), and the cost is calculated using the Mean Squared Error (MSE) over \( N \) samples:

\begin{equation}
\text{MSE} = \frac{1}{N} \sum_{i=1}^N \left( P^{RS}_i - R^{RS}_i \right)^2
\end{equation}

\begin{table}[htb]
    \centering
    \small
    \caption{Redshift Prediction Using Various Architectures with AstroMAE \citep{astromae}.}
    \begin{tabular}{|M{1.8cm}|p{2.8cm}|p{1cm}|p{1cm}|}
        \hline
        \textbf{Training Type} & \textbf{Architecture} & \textbf{MSE} & \textbf{MAE} \\ \hline
        \multirow{6}{=}{\centering Supervised Training \\ (from scratch)} 
        & plain-ViT-magnitude & 0.00077 & 0.01871 \\ \cline{2-4}
        & pcm-ViT-magnitude & 0.00057 & 0.01604 \\ \cline{2-4}
        &  \cellcolor[HTML]{C3C3C3}\citet{henghes} &  \cellcolor[HTML]{C3C3C3}0.00058 &  \cellcolor[HTML]{C3C3C3}0.01568 \\ \cline{2-4}
        & plain-ViT & 0.00097 & 0.02123 \\ \cline{2-4}
        & pcm-ViT & 0.00063 & 0.01686 \\ \cline{2-4}
        & Inception-only & 0.00064 & 0.01705 \\ \hline
        \multirow{8}{=}{\centering Fine-Tuning} 
        & plain-ViT-magnitude & 0.00068 & 0.01740 \\ \cline{2-4}
        & pcm-ViT-magnitude & 0.00060 & 0.01655 \\ \cline{2-4}
        & plain-AstroMAE & 0.00056 & 0.01558 \\ \cline{2-4}
        & \cellcolor[HTML]{C3C3C3}\textbf{pcm-AstroMAE} & \cellcolor[HTML]{C3C3C3}\textbf{0.00053} & \cellcolor[HTML]{C3C3C3}\textbf{0.01520} \\ \cline{2-4}
        & plain-ViT & 0.00086 & 0.01970 \\ \cline{2-4}
        & pcm-ViT & 0.00084 & 0.01945 \\ \cline{2-4}
        & plain-ViT-inception & 0.00059 & 0.01622 \\ \cline{2-4}
        & pcm-ViT-inception & 0.00059 & 0.01601 \\ \hline
    \end{tabular}
    \label{tab:performance_results}
\end{table}

Table \ref{tab:performance_results} shows the performance of various architectures combining a pretrained encoder, magnitude block, and the inception model. Rows labeled ``from-scratch" denote models where both plain-ViT \citep{vit} and pcm-ViT \citep{pcm_vit} are initialized and trained entirely from scratch. In contrast, the fine-tuning section includes models where pcm-ViT and plain-ViT are pretrained and frozen during the fine-tuning.  All modules in Table \ref{tab:performance_results} adhere to the architecture shown in Figure \ref{fig:astromae_architecture}.

The AstroMAE encoder is pretrained on 80\% of the data, 10\% for validation, and the remaining 10\% for fine-tuning. For fine-tuning, this 10\% is further split into 70\% for training, 10\% for validation, and 20\% for testing. The results in Table \ref{tab:performance_results} are based on inference over the 20\% testing samples from the fine-tuning model. The results underscore the rationale behind AstroMAE’s architectural choices. 


Plain-ViT and pcm-ViT models exhibit better results compared to their from-scratch counterparts due to pretraining. Pcm-ViT outperforms plain-ViT in both cases, suggesting that self-attention alone lacks high-frequency information, which convolutional layers intuitively capture. The Inception-only model does not outperform pcm-ViT trained from scratch. This highlights that while high-frequency information captured by the inception module is beneficial, broader, low-frequency patterns are also essential for accurate redshift prediction. Similarly, pcm-ViT and plain-ViT achieve better results when combined with the inception model. So AstroMAE incorporates the Inception model, enhancing the model’s ability to capture detailed features.

Moreover, the image data alone may be insufficient for optimal redshift predictions. A comparison between pcm-ViT and plain-ViT inception models with the proposed AstroMAE incorporating an additional magnitude block, as shown in Figure \ref{fig:astromae_architecture}, supports this insight. So magnitude values for each image band are integrated during fine-tuning. \textbf{Given the superior performance of pcm-AstroMAE, it has been selected for this study}.
 

\subsection{Proposed Framework: CAI}

Analyzing large astronomical data requires advanced distributed systems to handle concurrent jobs at scale. We propose a novel cloud architecture called Cloud-based
Astronomy Inference (CAI), using AWS Lambda functions \citep{serverless} to solve this challenge with significant speed-up improvement. Figure \ref{fig:cai_framework} shows an overview of the proposed framework. It has the following steps:

\begin{figure}[!htb]
    \centering
    \includegraphics[width=0.4\textwidth]{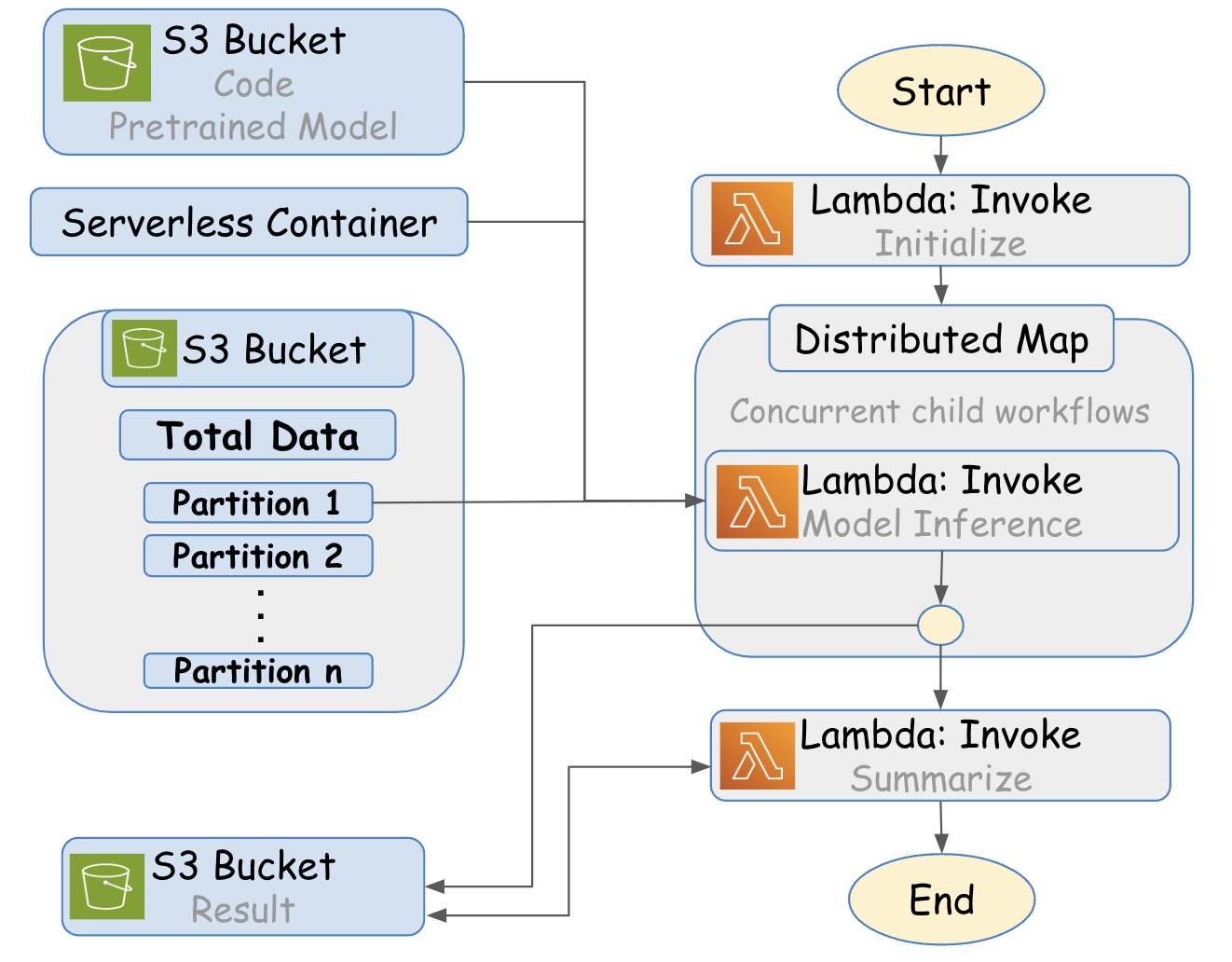}
    \caption{CAI framework overview using AWS Lambda Functions. It uses an AWS S3 bucket for data, code, and result storage. The state machine defines the workflow execution steps using AWS Lambda functions and distributed maps. Parallel execution is achieved through data partitions for almost linear high-performance inference scaling.}
    \label{fig:cai_framework}
\end{figure}

\begin{figure}[!htb]
    \centering
    \includegraphics[width=0.4\textwidth]{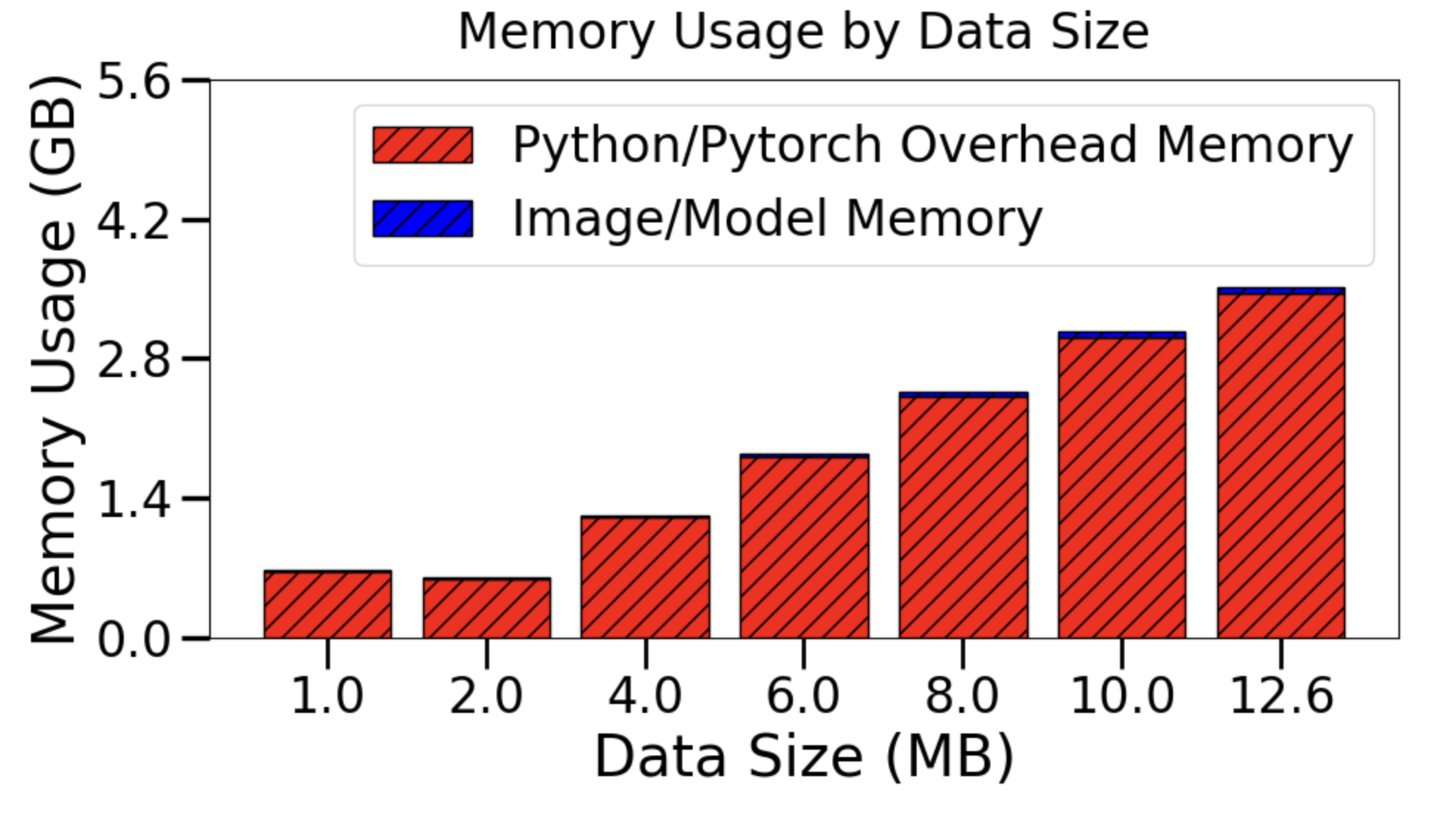}
    \caption{AWS Lambda Memory Usage by Partition Data Size.  We empirically size the dataset  based on the partition data size in MB.}
    \label{fig:memory-config}
\end{figure}

\begin{enumerate}
    \item \textbf{Initialization:} Defines the parameters and configurations of each child workflow based on the input payload. Then, it returns the parameter array with the state output. The concurrent job array size is the image data size divided by the smallest partition size. E.g. if the data and partition size is 1GB and 25MB respectively, the number of partitions is $\left\lceil 1024 \div 25 \right\rceil = 41 $.
    \item \textbf{Data Partitioning}: The total data is split into smaller sizes so that each lambda function can work with one small partition at a time during distributed processing. We use a maximum of 25MB of data per partition (arbitrarily chosen. Section \ref{sec:CAI_results} shows results for different partition sizes. This keeps the data loading time to a minimum and allows batch data processing. This is important because too large data can potentially run out of memory (AWS Lambda has a maximum of 10,240 MB memory). To correctly size the experiments, we empirically executed a series of experiments to determine the optimal image data size for processing, based on the maximum lambda size.  Based on this analysis, Figure \ref{fig:memory-config} illustrates the dataset and model memory usage, considering the partition data size in MB. However, this approach has a limitation compared to other serverless solutions like AWS Fargate (Serverless Docker on AWS ECS) or Apache OpenWhisk. The upper bound of available memory is a constraint in this approach. In contrast, AWS Lambda functions impose upper bounds on CPU and memory availability. As mentioned earlier, the memory limit for AWS Lambda functions is 10,240 MB. The CPU scaling and resource allocation of an AWS Lambda function are directly proportional to its memory configuration.
    
    \item \textbf{Distributed Processing:} This step is where concurrent job processing is performed based on each input item from the initialization. It uses the following AWS components: \begin{itemize}
        \item \textit{S3 Bucket:} Storage for code, the pre-trained model, and data. Each child process loads the code, model, and partitioned data (assigned by the initialization state).  We use the Python Boto3 library to interact and interface with our AWS S3 Bucket.
        \item \textit{Serverless Container:} The container is a custom AWS Lambda runtime created that includes all required software dependencies. The AWS Lambda function performing model inference is executed inside this runtime.  
        \item \textit{Distributed Map:} The distributed map state iteratively executes its child jobs based on the input array. We can control how many jobs run concurrently. Each Lambda job has a maximum of 10GB of RAM available (AWS limitation). \textbf{The model inference is done in this step independently for each partition and runs on CPU.} 
    \end{itemize}
    \item \textbf{Summarize Results:} The results from each job are summarized by this lambda function for final evaluation.
\end{enumerate}

\section{Experiments \label{sec:experiment}}
We intend to showcase how CAI is a novel approach for computing redshift prediction compared to other devices. This section thoroughly outlines the dataset used in our experiment, the metrics we use to assess each device, the experimental setup, and the analysis of our results.

\subsection{Dataset}

The dataset used in this study is prepared following \citet{astromae} and originates from \citet{pasquet}. It has \textbf{659,857 galaxy images derived from SDSS DR8} \citep{sdss_dr8}. Each image has dimensions of 64$\times$64$\times$5 and is annotated with 64 physical properties, including class, metallicity, and age, along with unique IDs for cross-referencing with other SDSS physical property tables. These images were captured using a 2.5-meter dedicated telescope at Apache Point Observatory in New Mexico and were already background-subtracted and photometrically calibrated. \citet{pasquet} further processed the images by resampling them onto a common grid of overlapping frames and applying the Lánczos-3 resampling kernel \citep{Lanczos} to enhance image quality and reduce artifacts. The images are center-cropped to a final size of 32$\times$32$\times$5, and the magnitude values for each band, along with the redshift, are obtained using the Astroquery library \citep{astroquery}. The resulting dataset, consisting of images and magnitudes as inputs and redshift as the target, is then prepared for model training.

\begin{figure}[!htp]
\centering
\includegraphics[width=.47\textwidth]{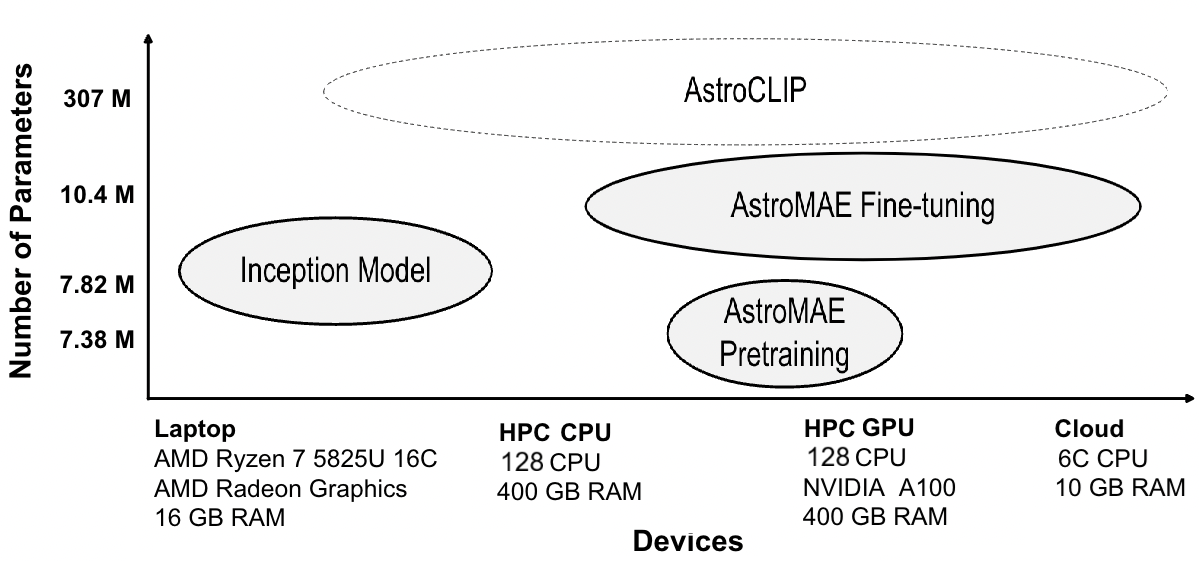}
    \caption{The parameter counts for recent deep learning-based methods developed for astronomy images, capable of inference across diverse computing environments—including a personal laptop, HPC CPUs, HPC GPUs, and our proposed cloud-based framework, CAI. A pre-trained AstroMAE model is used for the inference scaling experiments.}
    \label{fig:model_parameters}
\end{figure}

\subsection{Implementation
\label{sec:implementation}}

We used Python 3.11 with PyTorch 2 as the core framework. Also, NumPy 1.2 and Pandas 2.2 for data analytics. Additionally, Timm 0.4.12 was independently installed to provide access to specific model architectures compatible with PyTorch \citep{paszke2019pytorch}. The FMI and Boto3 libraries were used on AWS for the CAI implementation.

\subsection{AstroMAE model}
AstroMAE provides two pretrained models: the first is pretrained using 80\% of the data, as discussed earlier, and the second, which is employed in this study, is pretrained using the entire dataset. Both pretraining and fine-tuning are conducted on four A100 GPUs. \textbf{Pretraining takes approximately three days while fine-tuning on the full dataset requires around 10 hours}.

\subsection{Evaluation Metrics\label{sec:evaluation}}
To determine the scalability of performing redshift inference on each device, we evaluate using the metrics outlined and defined in Table \ref{tab:evaluation_metrics}. 

\begin{table}[!ht]
    \centering
    \small
    \captionsetup{justification=centering}
    
    \begin{tabular}{|>{\centering\arraybackslash}p{1.9cm}|p{5.1cm}|} 
        \hline
        \multicolumn{1}{|c|}{\textbf{Evaluation Metric}} & \multicolumn{1}{c|}{\textbf{Description}}\\
        \hline
        Memory Capacity & Maximum Dataset Size that can complete the redshift inference (GB)\\
        \hline
        Inference Time & Total time to complete redshift inference in seconds (s)\\
        \hline
        Throughput & Bits of data transmitted per second (bps)\\  
        \hline
    \end{tabular}
    \caption{Evaluation Metrics}
    \label{tab:evaluation_metrics}
\end{table}

These metrics effectively measure scalability. Processors that can complete inference on the full dataset are strongly preferred to those that cannot due to their memory limitations. When working with large astronomy datasets, this is more crucial. Faster model inference allows researchers to conduct more analyses. Lastly, a higher throughput means more data can be processed in less time.  These metrics are interdependent, and an ideal scalable framework can complete the inference on the full dataset and maximize inference throughput while minimizing time. We measure inference time using the PyTorch Profiler.

\textbf{For CAI}, since it is running parallel jobs, we calculate the inference time as the time needed to finish the Distributed Model Inference state (Figure \ref{fig:cai_framework}) from the AWS step machine. The throughput is also calculated similarly, based on the total bits calculated by all jobs divided by the total time to finish this state.

\subsection{Experiment Setup
\label{sec:experiment_setup}}
We compare the following processors: a personal laptop,  HPC CPU, HPC GPU, and our proposed CAI. Figure \ref{fig:model_parameters} shows their specifications, and Table \ref{tab:device_price} lists the maximum dataset size they can handle. A batch size of 512 is used throughout the experiment. Each experiment combination is repeated three times, and the average metrics are reported.

\begin{table}[!hbp]
    \centering
    \small
   \captionsetup{justification=centering}
    
        \begin{tabular}{|>{\centering\arraybackslash}p{1.4cm}|>{\centering\arraybackslash}p{1.4cm}|>
        {\centering\arraybackslash}p{1cm}|>
        {\centering\arraybackslash}p{1cm}|>{\centering\arraybackslash}p{1cm}|} 
        \hline
        \textbf{Device} & \textbf{Maximum Data (GB)} & \textbf{\#Cores} & \textbf{RAM (GB)} & \textbf{\#Images} \\
        \hline
        Laptop & 8 & 16 & 16 & 418K \\
        \hline
        HPC GPU & 12.6 & 128 & 400 & 660K \\
        \hline
        HPC CPU & 12.6 & 128 & 400 & 660K \\
        \hline
        CAI & 12.6 & 6 & 10 & 660K \\
        \hline
    \end{tabular}
    \caption{Hardware details and maximum data processed for each device. The original full dataset is 12.6 GB. }
    \label{tab:device_price}
\end{table}

\textbf{First,} we investigate how varying dataset sizes affect inference time, benchmarking the scalability of each device. The complete dataset is partitioned into incremental sizes: 1 GB, 2 GB, 4 GB, 6 GB, 8 GB, 10 GB, and 12.6 GB (the full dataset). This enables us to evaluate the total inference time on various devices and determine the maximum dataset size each device can handle. \textbf{Second,} we examine the impact of batch size on throughput (bps) across different devices. Batch sizes of 32, 64, 128, 256, and 512 are tested using a fixed dataset size of 1 GB. \textbf{Finally,} we evaluate the choice of partition size for CAI, using 25MB, 50MB, 75MB, and 100MB. This shows how different partition size choices could impact the inference time and throughput.

\subsection{Performance Results and Analysis\label{sec:analysis}}


\begin{figure}[!htb]
\centering
\includegraphics[width=0.5\textwidth]{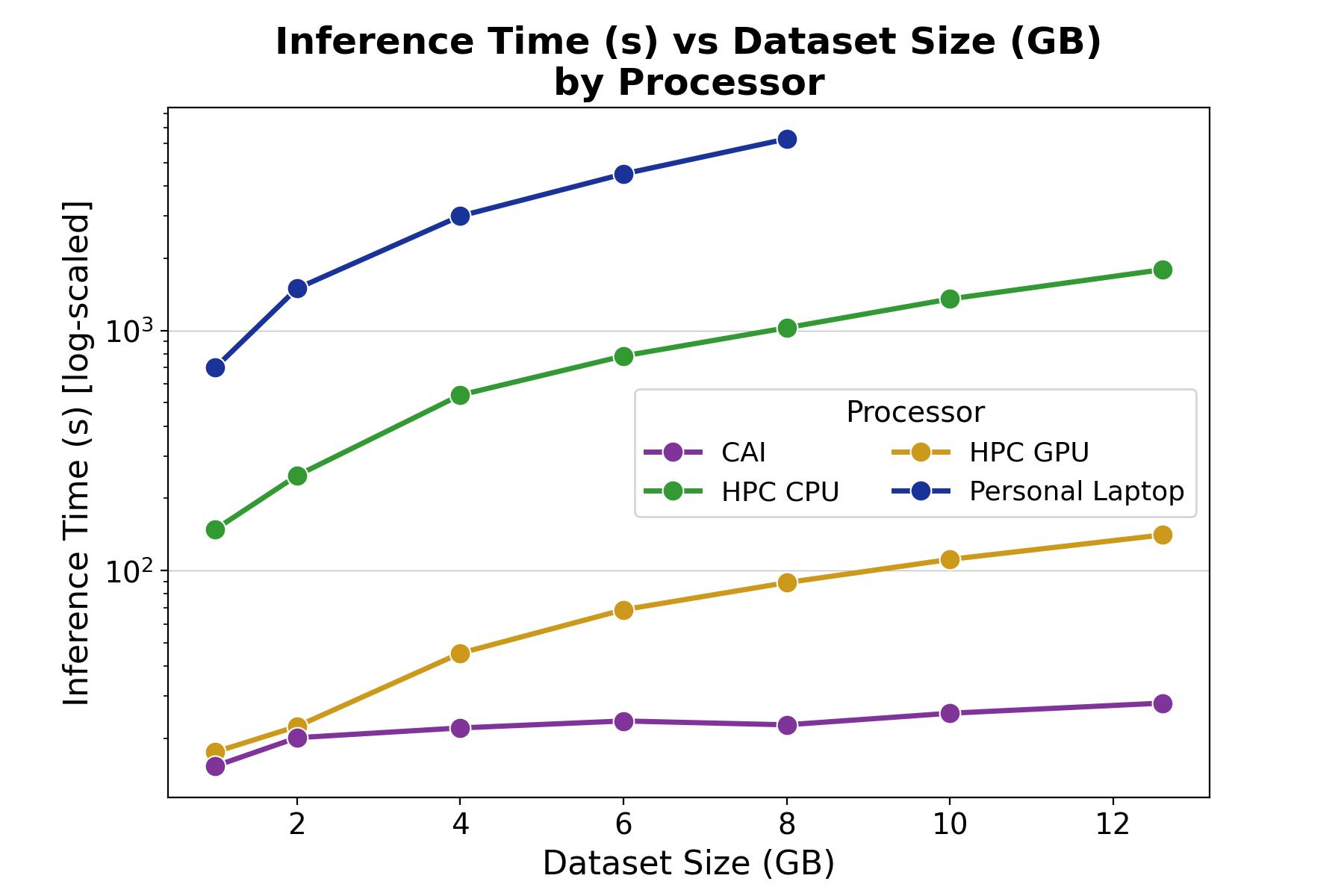}
\caption{Inference time vs dataset size by processor (batch size 512). Each combination is run three times and the average is reported. CAI performs better than an HPC GPU with its consistent scaling with increasing data size.}
\label{fig:time_gb}
\end{figure}

Figure \ref{fig:time_gb} shows that the inference time increases with data size. The personal laptop could not perform inference on the larger dataset sizes and is much slower due to memory limitations. The HPC with GPU is way faster than with a CPU and a personal laptop. However, CAI maintains a considerably stable inference time and performs best, due to its parallel executions, which is more evident with larger data sizes.

Model inference on the total data takes 1793 s, 140.8 s, and 28 s on the HPC CPU, HPC GPU, and CAI, respectively. The personal laptop could run with a maximum of 8GB and took an average of 6283 seconds. This showcases that personal devices are limited by memory and not feasible for the long run-times with astronomy data. Our proposed CAI is an attractive approach due to the faster inference speed despite running on a CPU only.

\begin{figure}[htbp]
\centering
\includegraphics[width=0.48\textwidth]{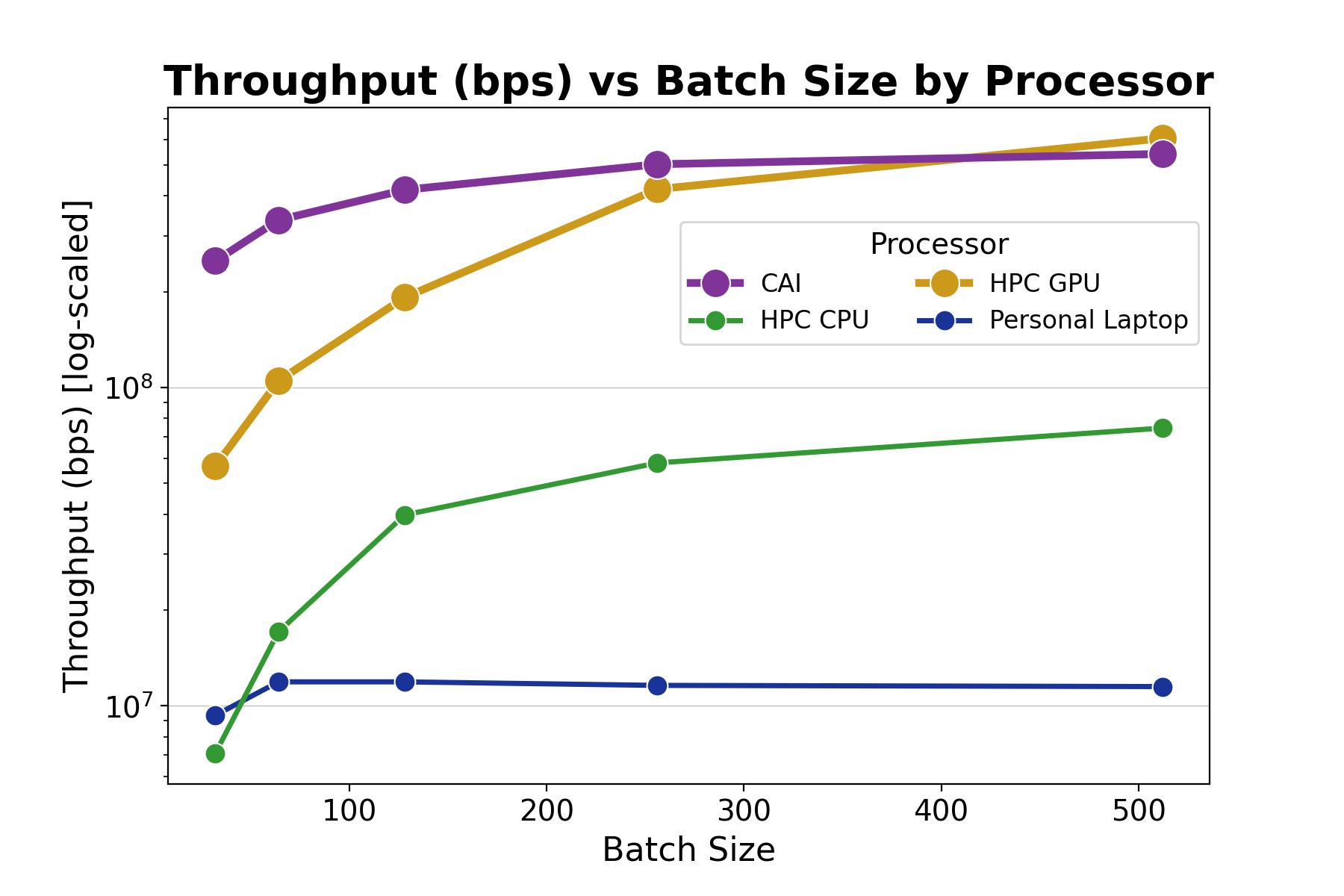}
\caption{Average throughput (bps) vs batch size by single node device to serverless computing using the 1GB dataset on a log scale. Three executions were completed for each batch size and processor combination, and the average throughput is displayed.}
\label{fig:throughput_batchsize}
\end{figure}

Figure \ref{fig:throughput_batchsize} shows the throughput of the devices for 1GB data size. 1GB was chosen due to the limitation of the personal laptop, taking significant time with a small batch size and larger data. Both the HPC GPU and CAI have significantly higher throughputs. CAI has the highest throughput (bps) for batch sizes 32, 64, and 128 (0.25B bps, 0.34B bps, and 0.42B bps, respectively). The HPC GPU has the highest throughput (bps) for batch sizes 256 and 512 (0.42B bps and 0.61B bps). After the CAI and the HPC GPU, the HPC CPU tends to have the next highest throughput, followed by the personal laptop. 

The trend of throughput (bps) by the processor is that throughput (bps) increases as batch size increases, and the throughput increase rate declines with increasing batch sizes. This is because there is an overhead computation cost at the beginning of iterating through each batch. The smaller batch sizes have more batches to iterate through, so although we might expect data to be processed quicker with smaller batch sizes, the increased frequency of the overhead computational cost can add up and lower the throughput. In fact, in Figure \ref{fig:throughput_batchsize}, CAI appears to plateau at the batch size of 256, and the personal laptop appears to plateau at the batch size of 64.


\subsection{Serverless Computing for Scalable Cosmic AI \label{sec:CAI_results}}

\begin{figure}[htb]
    \centering 
    \includegraphics[width=0.5\textwidth]{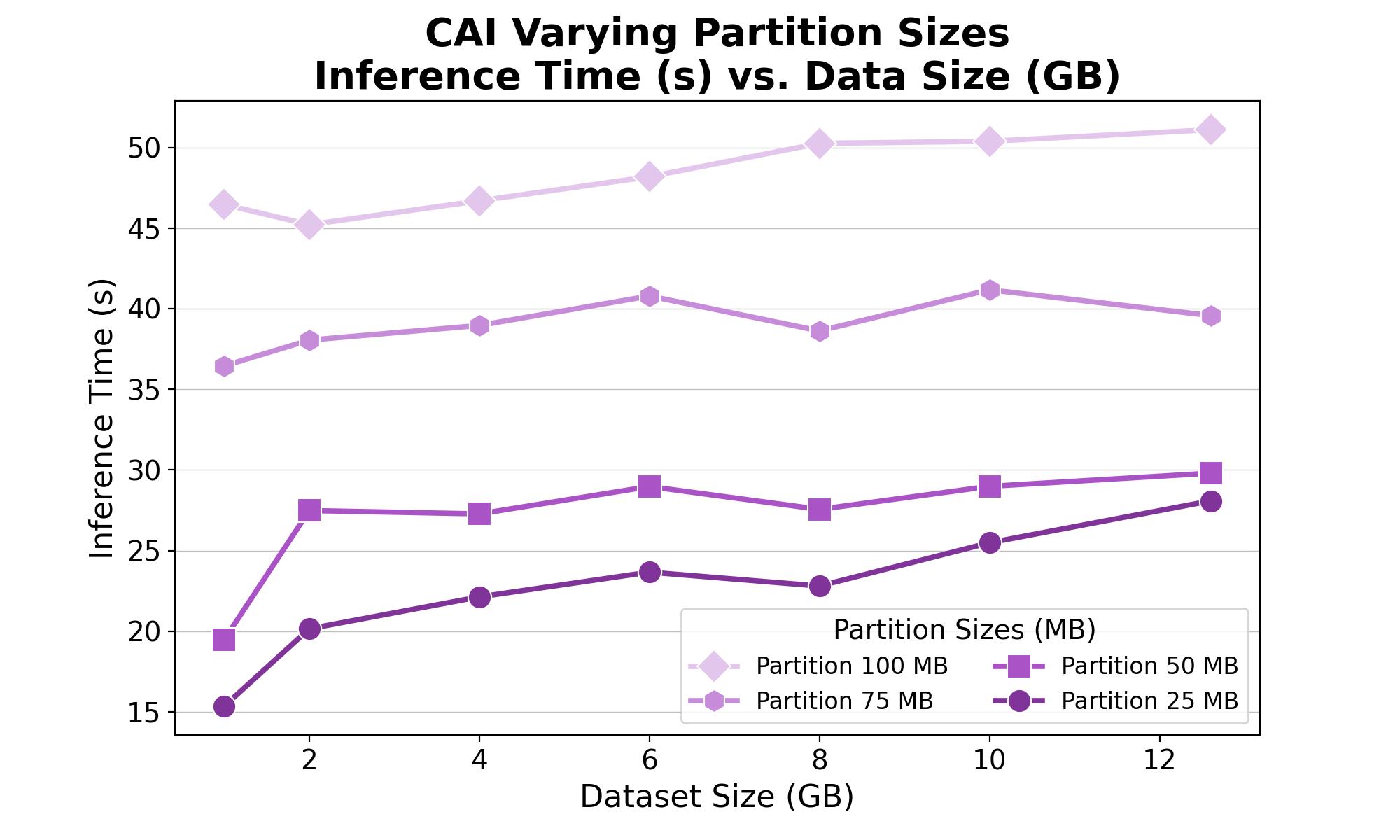}
    \caption{CAI average model inference time for different data sizes. The average time stays almost linear despite scaling up the data size due to concurrent jobs.}
    \label{fig:cai-inference-time}
\end{figure}

Figure \ref{fig:cai-inference-time} shows the CAI average model inference time in seconds for different data sizes. This complements Figure \ref{fig:time_gb} but shows the CAI performance in more detail. The inference time here is the runtime of the Model Inference State (Figure \ref{fig:cai_framework}), which completes only when all concurrent Lambda functions are done. This ensures we get the correct performance in practice. The individual Lambda jobs finish much faster, but that doesn't truly reflect the throughput.



\begin{figure}[!htb]
    \centering
    \includegraphics[width=0.5\textwidth]{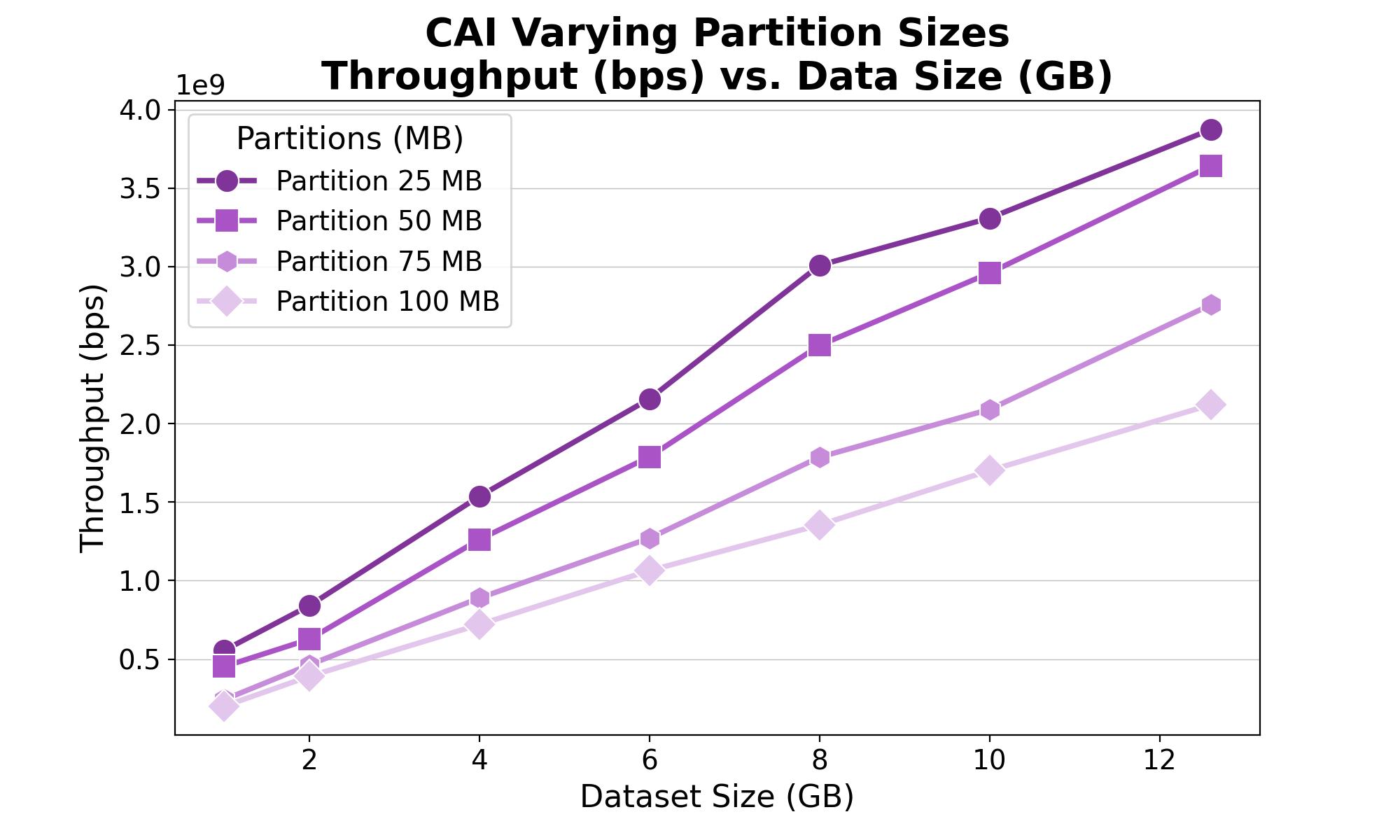}
    \caption{CAI Average Throughput for Different Data Sizes. These results are over concurrent jobs and use a batch size of 512 for different partition sizes.}
    \label{fig:cai-throughput}
\end{figure}

Despite increasing the dataset size, the inference time in CAI is almost constant. For example, for the 100 MB partition, the average inference time for the 1 GB dataset is 23.02 s and only increases by 1.56 s to 	24.58 s for the full 12.6 GB dataset. Furthermore, for the 25 MB partition, the average inference time for the 1 GB dataset is 6.56 s and subtly decreases by 0.73 s to 5.83 s for the full 12.6 GB dataset. This is due to the parallel processing of the large data into smaller partitions. Since each child workflow is running model inference on a small partition of data independently, the overall run time depends mostly on the execution time for that partition. 

Figure \ref{fig:cai-throughput} shows the throughput for those different partition sizes across data sizes. For each dataset size, the 25 MB partition has the highest throughput, then the 50 MB partition, then the 75 MB partition, and lastly, the 100 MB partition. At the full dataset size of 12.6 GB, the 25 MB partition has an average throughput of 18.04B bps, the 50 MB partition has an average throughput of 9.70B bps, the 75 MB partition has an average throughput of 6.66B bps, and the 100 MB partition has an average throughput of 5.12B bps. With a smaller partition, we can use more concurrent jobs, increasing the throughput. 

\begin{figure}[!htp]
\centering
\includegraphics[width=.47\textwidth]{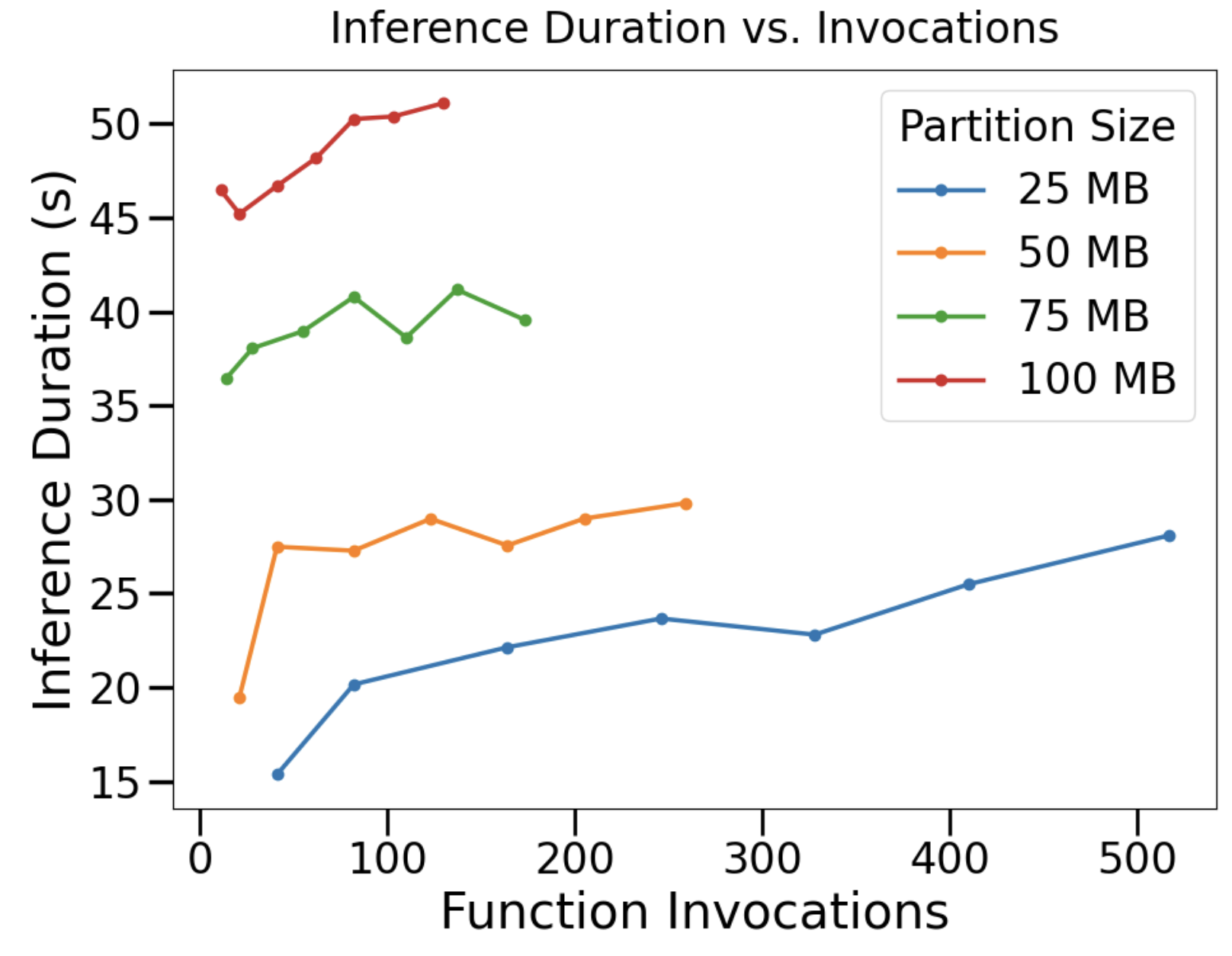}
    \caption{CAI: number of AWS Lambda Function invocations for different data sizes.  Similar to figure \ref{fig:cai-throughput}, results use a batch size of 512 and are displayed across partition sizes.}
    \label{fig:cai-invocations}
\end{figure}

With AWS Lambda, it is quite trivial to scale out horizontally to handle large dataset sizes.  Figure \ref{fig:cai-invocations} details the number of invocations and inference duration to illustrate scalability.  The number of invocations is determined by dividing the partition size by the dataset size.  


\textbf{Runtime Cost analysis:} AWS Lambda costs are based on Table \ref{tab:aws_published_costs}. For CAI, we use three AWS Lambda functions: 1) data-parallel-init-fmi, 2) cosmic-executor, and 3) resultSummary.  The material cost is the number of executions of the cosmic executor.  The costs for 1) and 2) are marginal based on the fact that these are only single executions for a given job (i.e., 25 GB, 50 GB, 75 GB, or 100 GB partition(s) and larger scaling tests for the 100GB partition for 256 GB, 512 GB, 768GB and one terabyte data sizes), and each of these lambda functions is provisioned very conservatively with 128MB memory and 512MB ephemeral storage. AWS published guidance indicates Lambda costs for the CAI use case to be based on architecture (x86 or ARM), memory, ephemeral storage, and data transfer for CAI.  We have provisioned our cosmic-executor AWS Lambda function as detailed in Table \ref{tab:aws_config}.

\begin{table}[htbp]
    \centering
    \small
    \begin{tabular}{|>{\centering\arraybackslash}p{4cm}|>{\centering\arraybackslash}p{3cm}|}
    \hline
    \textbf{Detail} & \textbf{Cost} \\ 
    \hline
    \textbf{x86 Arch}: First 6 Billion GB-seconds/month; \newline \textbf{Duration}: \$0.0000166667 per GB-second & \$0.20 per 1M requests \\
    \hline
    \textbf{Memory (MB)}: 10,240 & \$0.0000000021 per 1ms \\
    \hline
    \textbf{Ephemeral Storage} & \$0.0000000309 for every GB-second \\
    \hline
    \end{tabular}
    \caption{AWS Lambda Costs for CAI Executions}
    \label{tab:aws_published_costs}
\end{table}

\begin{table}[htbp]
    \centering
    \small
    \captionsetup{justification=centering}
    \begin{tabular}{|r|c|M{1.2cm}|c|M{1.2cm}|} \hline
         \textbf{Memory} & \textbf{Ephemeral} & \textbf{Timeout}  & \textbf{Snapshot} \\ \hline
         10240MB & 10240MB & 15min0sec & None  \\\hline
         
    \end{tabular}
    \caption{Data Parallel AWS Lambda configuration}
    \label{tab:aws_config}
\end{table}

The cost of invoking the CAI cosmic-executor AWS Lambda function is \$0.00001667 per GB-second of computation time. It should also be mentioned that we have small S3 costs related to storage (our S3 Bucket is around 50GB) and data transfer. 
 For the experiments conducted in the aggregate, this amounts to \$0.023 per GB, and data transfer costs \$0.005 per 1,000 requests.  This cost is non-material, similar to costs related to executing the data-parallel-init-fmi and resultSummary AWS Lambda function.  Figure \ref{fig:cai_framework} shows that our framework calls the Lambda function during initialization, parallel processing, and summarization. Table \ref{tab:aws_cost} summarizes some example cases to estimate the computation cost for our task. The number of requests represents the frequency at which the Lambda function was called, which is equivalent to the number of concurrent jobs (data divided by partition size). 

\begin{table}[htbp]
    \centering
    \small
    \begin{tabular}{|r|c|M{1.5cm}|c|M{1.2cm}|} \hline
         \textbf{Partition} & \textbf{Requests} & \textbf{Lambda Duration}  & \textbf{Memory} & \textbf{Cost (\$)}  \\ \hline
         25MB & 517 & 5.70 s & 2.8GB & 0.16 \\\hline
         50MB & 259 & 10.8 s & 4.0GB & 0.20 \\\hline
         75MB & 173 & 15.7 s & 5.9GB & 0.30 \\\hline
         100MB & 130 & 21.0 s & 7.0GB & 0.38 \\\hline
         100MB & 10240 & 23.0 s & 7.0GB & 27.48 \\\hline
    \end{tabular}
    \caption{An estimation of AWS computation cost for inference on the total dataset up to one terabyte data size. Cost in US cents is $requests \times duration(s) \times memory(GB) \times 0.00001667 $.}
    \label{tab:aws_cost}
\end{table}

All this amounts to under five US dollars to execute the experiments for all but the largest experiment.  One novelty aspect of CAI is related to the costs of executing astronomy inference compared to similar costs on HPC clusters or other "rack and stack infrastructure."  


\subsection{Large Scale Experiment and Analysis}

We demonstrate the scalability of our CAI framework by conducting experiments on significantly larger datasets, up to 1 Terabyte (1024 GB). This is achieved by repeatedly sampling from the original 12.6 GB data partitions. We then compare CAI's performance with that of a traditional High-Performance Computing (HPC) environment. The experiments were performed on HPC CPUs to make a fair comparison with CAI, which is also CPU-only. Both setup uses the 100 MB file partition for consistency. Table \ref{tab:extended_setup} shows the function invocations for each data size. This is determined by the number of partitions we need to process to reach the target data size. Calculated by dividing the data size in MB by the partition size, 100 MB. 


\subsubsection{HPC CPU Setup:} The HPC environment uses the SLURM job scheduler \citep{slurm} to submit a large number of concurrent jobs. Similar to CAI, each job processes a single file partition on a single CPU core. The `srun` Python command then executes the inference on all allocated CPU cores. The concurrency limit for CPU cores on the HPC was 6000; beyond that, it'll wait for the running jobs to finish. The virtual environment uses the same Python library versions as the AWS CAI framework. Each HPC node had maximum 96 cores and 750 GB of memory. 

\begin{table}[!htp]
    \centering
    \small
    \begin{tabular}{|c|c|c|c|c|c|} \hline
        \textbf{Data (GB)} & 100 & 256 & 512 & 768 & 1024 \\ \hline
        \textbf{\#Parallel Functions} & 103 & 2622 & 5243 & 7865 & 10486\\ \hline
    \end{tabular}
    \caption{Setup summary of the extended scaling analysis: run up to 1TB data over 10486 parallel CAI Lambda functions on AWS and 10486 parallel functions on the HPC system.}
    \label{tab:extended_setup}
\end{table}

\subsubsection{Results and Discussion:}

\begin{figure}[!ht]
    \centering
    \includegraphics[width=.48\textwidth]{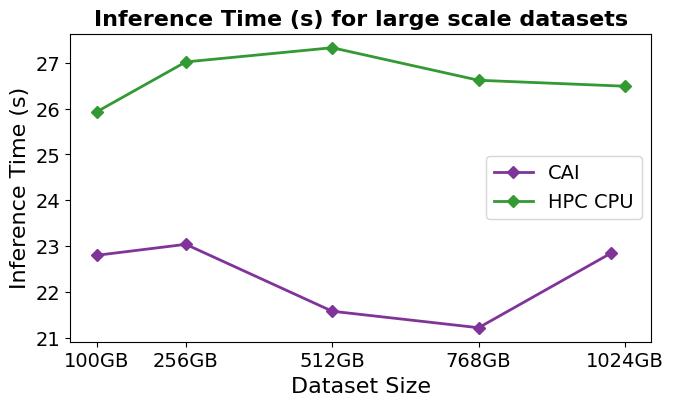}
    \caption{Extended large-scale scaling experiment: Comparing CAI Lambda and the HPC system on inference time with up to 1 TB of data and 10,486 parallel functions.}
    \label{fig:cai_hpc_extended}
\end{figure}

Figure \ref{fig:cai_hpc_extended} shows the average inference time for both the CAI framework and the HPC CPU environment across various large datasets, from 100 GB to 1024 GB. The results clearly demonstrate the superior scalability of the CAI framework.

CAI maintains a consistently low inference duration across the entire range of dataset sizes. This linear scalability is a direct result of its serverless architecture, which allows a high degree of parallelization by processing multiple data partitions concurrently. The inference time is primarily dependent on the time it takes to process a single data partition, not the total dataset size. 

\section{Conclusion and Future Work}

This paper presents a novel cloud-based framework, CAI, designed to enhance the inference scalability of foundation models trained on astronomical images. Our study highlights the essential role of cloud services in overcoming the computational and storage challenges inherent in modern astronomical research. We have leveraged serverless computing to enhance the scalability of foundation models via partitions and data parallel optimizations.

To showcase the attractive capabilities of CAI, we used a recent foundation model, fine-tuned for redshift prediction, in our experiments. Comprehensive evaluations across varying dataset sizes and computing devices demonstrate CAI’s robustness in scaling almost linearly on the inference and high throughput for large-scale astronomical image datasets. Note that this framework can be applied to additional critical astronomy inference tasks, such as morphology classification and star formation history. Future work will also involve testing other foundation models developed for astronomical images and spectra. 

In the future, we plan to introduce integration with \textbf{FMI, FaaS Message Interface} \citep{faas1} for high-performance communication between AWS Lambda functions and to introduce support for collective operations. Direct communication between AWS Lambda functions over TCP sockets is not supported, based on published capabilities and the underlying design of Lambda functions by Amazon, which are intended to process lightweight event notifications or request payloads.  To elaborate, AWS Lambda communication is capped at 6 MB for synchronous request/response, 20 MB for streamed data, and 256 KB for asynchronous.  Direct communication offers performance benefits related to message passing between Lambda functions compared to JSON payloads and other solutions that introduce intermediary data storage to overcome communication performance and size limitations.  With the FMI integration, functions will be able to participate in point-to-point, high-performance message passing and perform collective operations during later stages (e.g., model inference). For our scaling experiments with data sizes of 256, 512, 768, and one terabyte, it was necessary to use intermediate storage based on inbound and outbound payload size limitations. FMI \cite{faas2} demonstrates significant improvements ranging from 105 to 1025 times compared to storage mechanisms like AWS S3, Redis, or AWS DynamoDB, we can enhance both costs and performance. Furthermore, we can tackle the crucial challenge of communication performance in AI/ML inference on AWS Lambda. 

\section{Acknowledgement}
This work is partly supported by NSF-Simons AI Institute for Cosmic Origins (CosmicAI: Grant 2421782) Seed Grant. NSF CF-1918626 Expeditions: Collaborative Research: Global Pervasive Computational Epidemiology, and NSF Grant 2200409 for CyberTraining:CIC: CyberTraining for Students and Technologies from Generation Z.

\bibliographystyle{SageH}
\bibliography{bibliography}
\end{document}